\documentclass[sigconf,nonacm]{aamas}

\usepackage{balance} 

\setcopyright{ifaamas}
\acmConference[AAMAS '25]{Proc.\@ of the 24th International Conference
on Autonomous Agents and Multiagent Systems (AAMAS 2025)}{May 19 -- 23, 2025}
{Detroit, Michigan, USA}{A.~El~Fallah~Seghrouchni, Y.~Vorobeychik, S.~Das, A.~Nowe (eds.)}
\copyrightyear{2025}
\acmYear{2025}
\acmDOI{}
\acmPrice{}
\acmISBN{}

\acmSubmissionID{25}

\title[MA DRL for Ramp Entry]{A Systematic Study of Multi-Agent Deep Reinforcement Learning for Safe and Robust Autonomous Highway Ramp Entry}

\author{Larry Schester}
\affiliation{
  \institution{Flex} 
  \city{Farmington Hills, Michigan}
  \country{United States of America}}
\email{larryschester@gmail.com}

\author{Luis E. Ortiz}
\affiliation{
  \institution{University of Michigan-Dearborn}
  \city{Dearborn, Michigan}
  \country{United States of America}}
\email{leortiz@umich.edu}

\begin{abstract}
Vehicles today can drive themselves on highways and driverless robotaxis operate in major cities, with more sophisticated levels of autonomous driving expected to be available and become more common in the future. Yet, technically speaking, so-called "Level 5" (L5) operation, corresponding to full autonomy, has not been achieved. For that to happen, functions such as fully autonomous highway ramp entry must be available, and provide provably safe, and reliably robust behavior to enable full autonomy. We present a systematic study of a highway ramp function that controls the vehicles forward-moving actions to minimize collisions with the stream of highway traffic into which a merging (ego) vehicle enters. We take a game-theoretic multi-agent (MA) approach to this problem and study the use of controllers based on deep reinforcement learning (DRL). The virtual environment of the MA DRL uses self-play with simulated data where merging vehicles safely learn to control longitudinal position during a taper-type merge. The work presented in this paper extends existing work by studying the interaction of more than two vehicles (agents) and does so by systematically expanding the road scene with additional traffic and ego vehicles. While previous work on the two-vehicle setting established that collision-free controllers are theoretically impossible in fully decentralized, non-coordinated environments, we empirically show that controllers learned using our approach are nearly ideal when measured against idealized optimal controllers.
\end{abstract}


\newcommand{\BibTeX}{\rm B\kern-.05em{\sc i\kern-.025em b}\kern-.08em\TeX}

\begin{document}


\pagestyle{fancy}
\fancyhead{}


\maketitle 



\section{Introduction}

The popular view is that highly \emph{autonomous vehicles (AVs)} are a desirable future technology, expected to save lives and increase mobility. The evolution of this technology involves mixed-usage scenarios with roadway sharing between human-operated and AVs of varying capabilities. At least for the foreseeable future, effective, safe road sharing requires AVs to act or behave like those operated by human drivers. Ideally, AVs must perform maneuvers as well as or better than humans and achieve optimal performance with minimal collisions.


\begin{figure}
  \centering
  \includegraphics[width=\linewidth]{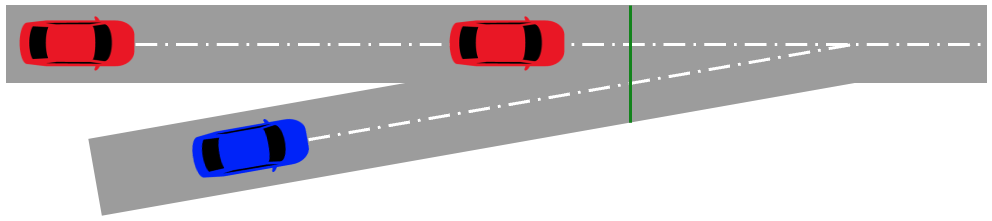}
  \caption{{\bf Three-Vehicle Highway Merge}: Goal line is green. Ego merge vehicle (blue) and two traffic vehicles (red).}
  \Description{Three-vehicle highway merge scene with a tapered on-ramp.}
  \label{fig:3vehsim}
\end{figure}

\begin{figure}
  \centering
  \includegraphics[width=\linewidth]{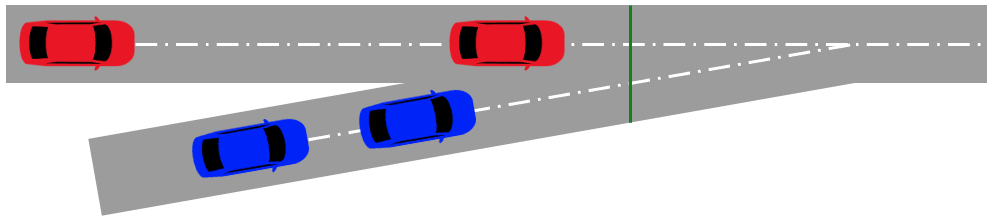}
  \caption{{\bf Full-Scene Highway Merge}: Two vehicles in the merge lane and two or more vehicles in the traffic lane.}
  \Description{Four-vehicle highway merge scene with a tapered on-ramp.}
  \label{fig:4vehsim}
\end{figure}

Most highway driving consists of maintaining speed and slowing to maintain a reasonable gap to vehicles in front. This function is handled today by widely available \emph{adaptive cruise control (ACC)} systems. Maintaining position within a lane is performed with a lane keeping/centering function. Most highways have at least two lanes, one of which is for passing, so lane changing is another function needed for full autonomy. These systems can combine with common navigation systems and in-vehicle traffic sign recognition cameras to get the vehicle to a specific destination. Special functions like traffic jam assist also exist. Together these functions could navigate through interchanges, construction zones, or other aspects of highway driving. Most L2-L3 systems commercially available today provide at least ACC, lane centering, and lane changing working together to automate driving, but require the driver to maintain constant vigilance and always understand the road scene in order to take over from the system with or without notice. Examples are GM Supercruise ~\citep{GMSupercruise}, Ford Bluecruise ~\citep{FordBluecruise}, and Tesla Full Self Driving ~\citep{TeslaFSD}. 

Our focus is on L4 or L5 \emph{highly autonomous driving (HAD)}. Here, completely driverless, robust, and verifiable systems must be implemented because highway driving has an innately greater risk due to the higher resulting forces of high-speed travel and faster reaction times needed. A deep understanding and verification of these systems that enable HAD is necessary and essential for their success. Our work seeks to add to the body of knowledge required to eventually make these systems a reality.

Autonomous or \emph{Advanced Driver Assistance Systems (ADAS)} equipped vehicles exist today in two major forms: L4 robotaxis and L2-L3 assistance (as mentioned above). No L5 systems exist today and most experts generally do not expect them anytime soon because solving the problem is challenging.
Most L4 vehicles drive in limited, highly-mapped geo-fenced areas of cities where the same routes are travelled repeatedly. Speeds are usually slower than highway speeds and where stopping all motion when an unknown situation arises is a generally acceptable safe response.
A fully automated merge maneuver is necessary for HAD L4-L5 as defined by the SAE International J3016 standard ~\citep{SAEInternational2020}, a well-known body within the automotive industry. The challenges of merging are relatively straightforward for humans to understand. To successfully merge into traffic, a driver must quickly assess the road scene to determine the best way to control the vehicle to insert it into a traffic stream while judging and reacting to the behavior of the other vehicles, all while minimizing the risk of a  collision. It is a complex task for humans to learn and no less complex to automate. 

In "crowded" environments, like those in most highways near metropolitan areas, merging is a very risky complex maneuver that most humans learn to navigate reasonably well with sufficient practice/experience. However, we must keep in mind the precise level of complexity of the interactions involved that any autonomous system must reason about as it learns to navigate it: the driver does not know if the other traffic is going to be let in, block, or ignored. Any driver must decide how to speed up or slow down all within a few seconds or less while traveling down a merge ramp that pushes you into a stream of traffic. Considering this, the level of complexity of merging is much greater than the other functions above that make up fully autonomous highway driving, like ACC or even lane changing. In most highway driving, lane changing is generally not necessary or time-constrained in non-emergency situations, but ramp merging is. 

\emph{In this paper, we systematically study the application of a multi-agent (MA) approach based on game theory and deep reinforcement learning (DRL) to the challenging problem of autonomous vehicle taper-type highway-ramp merging involving more than two vehicles in the road scene (Fig.~\ref{fig:3vehsim}). We empirically show via simulation that the controllers learned via self-play training using our approach are near optimal during test performance.}

\section{Autonomy Challenges}

The general societal view is that AVs need to be safer and less fallible than humans. However, this seems to be a difficult goal to obtain or even measure. 
Recently we have seen increased scrutiny in the industry as it has come to a reckoning
about this difficulty despite incredible investment~\citep{bloomberg2022}.
As a result, some major companies have pulled development funding and shuttered technology partnerships focused on the efforts~\citep{Argo2023, Motional2024}.

Recent incidents with AVs and pseudo-AVs~\citep{Boudette2021,tusimple2022} are leading to increased mistrust from the public and calls for greater regulation of the technology~\citep{koopman2024redefining,cummings2024identifying}. Questionable safety~\citep{cruise2024}, unexpected reliance on remote human operation~\citep{CruiseAssistance2023}, and subsequent under-reporting to government agencies~\citep{CruiseShutdown2023} have placed a greater spotlight on the viability of true autonomy.

When accidents do occur, there is a trend for companies to place blame on the driver since most technology available to drivers today is labeled as ADAS (i.e., L2 or below), which requires the driver to be constantly vigilant and in full control of the vehicle despite the drivers impression that the vehicle may be driving itself~\citep{Penmetsa2019,TeslaLawsuit2024}.
Ultimately, safe use of these systems could benefit from an advanced licensing options to ensure drivers are capable of using the systems, just like licensing for drivers of commercial vehicles. 

Industry hype can make people believe that full autonomy has arrived en masse, but this is not reality ~\citep{Boudette2021}. The state of the art today for \emph{artificial intelligence (AI)} within AVs can at times have surprisingly rudimentary problems. For example a recent news story has reported that L4 capable Waymo vehicles seeking to arrange themselves within a parking lot overnight have troubles coordinating their behavior, resulting in AVs honking at each other in an effort to avoid collisions while unfortunately keeping nearby residents awake ~\citep{vergeWaymoHonking2024}. Cruise had similar difficulties where vehicles traffic jams were caused when multiple vehicles have inexplicably stopped at the same intersection, needing to ultimately be cleared by human drivers ~\citep{cruiseJam2023}. While the perception may be that these L4 vehicles are operating fully autonomously since there is no driver, the reality is that most of these vehicles rely frequently upon remote drivers ~\citep{NYtimesRemoteDriver2024}. China, the leader in driverless testing according to~\citet{ChinaDriverlessLeader2024}, has even enacted legislation about the number of remote drivers needed to support so-called 
full AVs~\citep{chinaRemote2024}.

Many companies today rely upon brute force testing where millions of miles are accumulated, implying that this will be sufficient to prove safety~\citep{Cruise2023, Victor}. However, accumulation on the order of billions of miles may be needed~\citep{Koopman2022}. Impeccably safe operation would also be needed in far greater scenarios of operation, like dense and sparse traffic, city and rural areas, fair and inclement weather, day and night driving, and all conditions in between. Digital twins and other simulation techniques help test more treacherous scenarios faster while maintaining safety~\citep{Feng2023}. However, there are a lack of standards and test methodologies, unlike those already in time-tested existence for other safety technologies well-known in the automotive industry like braking, steering, and crashworthiness from Federal Motor Vehicle Safety Standards (FMVSS), National Highway Traffic Safety Administration's (NHTSA) New Car Assessment Program (NCAP), Insurance Institute for Highway Safety (IIHS), etc. 

However, the path to increased adaption of these technologies and the road to true autonomy remain unclear. What is clear is that safety remains paramount in the development and use of these systems to protect drivers, passengers, and the public from unnecessary harm. To ensure the current state of the art technology functions in an expected way, AD technology needs to be robustly vetted through validation and testing. Our approach uses a framework to test and analyze AD highway lane merging functions to help ensure safety more robustly and reliably. 

\section{Related Work}

There has been increased interest on the problem of AV merging over the last half-a-decade. 
Not surprisingly, \emph{reinforcement learning (RL)} approaches are growing in popularity as a way to obtain optimal controllers for the merging problem from experience given the evident success of RL in a large variety of control-related problems and domains. In particular, to handle the continuous and highly dimensional nature of the physical model governing the possible driving choices and the state of the interactions between vehicles while merging, 
DRL is applied as a way to efficiently represent (approximately optimal) policies and their quality using deep (artificial) \emph{neural-network (NN)} architectures. Given the natural interaction between highway on-ramp and in-traffic vehicles, 
MA 
approaches are of particular interest to us. For example, 
\citet{Shalev-Shwartz2016} present a DRL method that uses an option graph similar to \emph{long short-term memory (LSTM)} NN approaches. Their option graph considers the changing environment as a MA approach instead of a simplified single-agent \emph{Markov Decision Process (MDP)} approach. While they consider it an MA approach, it differs from the approaches used in our work. 
More recently, \citet{Kamran2021} formulate a cooperative merging scenario as a partially observable MDP 
for a trajectory planner and use 
deep-Q networks 
to solve it. 

Our work here builds on that we previously presented in~\citet{Schester2021}, and includes work already included in~\citet{schester_phd}.
They also use a game-theoretic MA-DRL approach in which speed changes (i.e., acceleration) control the longitudinal position of the ego vehicle agent in the merge lane to try to avoid a collision with the traffic vehicle agent as they reach the merge point. Also as in our work here, simulated data trains the DRL network for the merging vehicle. The data is a range of merge scene parameters and three different action behavior policies: constant, random, and reactive acceleration of the traffic agents. Finally, the same three policies test the trained agent, but with some parameters held constant using a structured, standardized test framework specifically developed to measure automated merging performance.  \citet{Schester2021} show decent success in obtaining controllers that avoid collisions during merging, but exclusively considers a limited two-vehicle scene. Hence, that work is not yet fully representative of a real highway merge because it is missing vehicles that could be present in general merging interactions. Our work here progressively adds additional vehicles to the scene using a similar DRL approach and a mixture of training and testing policies (Fig.~\ref{fig:3vehsim}). We first add an additional vehicle in the traffic lane (Fig.~\ref{fig:4vehsim}). Finally, we add another vehicle in the merge lane and enable a full stream of traffic for the merging vehicles to contend with (Fig.~\ref{fig:3vehsim}). Our work shows near optimal performance continues through scaling up the representative model.

\citet{Tang2019} presents a MA-DRL approach to merging. Their approach is MA in the sense that there are multiple intelligent agents interacting in a common environment, similar to the sub-combination we present with reactive traffic. 
They too use self-play simulation for training. They create a mix of rule-based and RL-trained agents in their simulation. They use a $76$-d vector that includes the position, velocity, acceleration, orientation, and turn-signal state of eight neighboring vehicles and input it into a \emph{convolutional neural-network (CNN)}-DRL model to generate three actions: steering, acceleration, and turn signal. In contrast, our research does not consider steering or turn signals because it focuses on the ego vehicle's longitudinal positioning to try to avoid collisions. In addition, their study focuses on a different merge scenario called a ‘zipper merge’ that is like a parallel-type lane merge, but the end of the merge lane continues into another ramp. The scenario is more like a lane change problem with a fixed distance to change lanes versus the taper-type merge ramp of our work.


There is also an active line of research on lane merging that also includes work in ramp merging with approaches based on RL~\citep{Triest2020}, intention estimation~\citep{Dong2018}, probabilistic graphical model (PGM)~\cite{dong2017interactive}, and prediction and cost functions~\citep{Wei2013}. 
That work focuses on the in-lane traffic vehicle as it approaches the merge intersection, unlike the work presented here that focuses on the ego vehicle as the one merging into traffic from the ramp. They also use Next Generation Simulation (NGSIM) datasets to train and evaluate their methods. However, the work presented here makes limited use of standard datasets because of the reactive nature of merging.

\begin{table*} 
\caption{State Parameters for both Three-Vehicle ($3V$) \& Full-Scene ($FS$) Highway Merge Models. Merge and traffic vehicles use different sets of state variables where \emph{Yes} and \emph{No} values indicate state variable usage per vehicle. When the information between the models differs, the \emph{full-scene}  version is added \emph{parenthetically (emphasized)}, or indicated with $3V$ \& $FS$ labels.}
    \centering
    \begin{tabular}{lcccc}
    \toprule
    \textbf{Vehicle state variables} & \textbf{Merge} & \textbf{Traffic} & \textbf{Range} & \textbf{Units} \\
    \midrule
    Closing gap to (\emph{rear}) vehicle in next lane & Yes & Yes & $[-2.5,30]$ & \emph{meters (m)} \\
    Relative closing speed to next lane (\emph{rear}) vehicle & Yes & Yes & $[-10,10]$ & \emph{meters/second (m/s)}\\
    Closing gap to (\emph{next lane}) front traffic & Yes & No (\emph{Yes}) & $[-2.5,30]$ & $m$ \\
    Relative closing speed to (\emph{next lane}) front traffic & Yes & No (\emph{Yes}) & $[-10, 10]$ & $m/s$ \\
    Closing gap to goal & Yes & No & $[-160, 150]$ & $m$ \\
    Closing velocity to goal & Yes & No & $[0, 40]$ & $m/s$ \\
    Closing time from current to front vehicle (TIV) & No (\emph{Yes}) & Yes & $[0, 2.5]$ & $s$ \\
    Time to goal position & No &Yes & $[0, 3]$ & $s$ \\
    Proximity to ego ([$3V$] behind or in front of; [$FS$] twice: one front, one rear) & No & Yes & $\{-1, 1\}$ & unitless \\
    \bottomrule
    \end{tabular}
    \label{tab:States}
\end{table*}

\section{Three-Vehicle Simulation}

The Three-Vehicle Simulation (Fig.~\ref{fig:3vehsim}) consists of one vehicle in the merge lane and two in the traffic lane. The ego merge vehicle must now contend with front and rear traffic vehicles and avoid collisions. This three-vehicle model is able to avoid collisions well under the 
$256m$ ramp length recommended by the U.S. DOT (Department of Transportation) \citep{AchmadAliFikriSyamsulArifin2022}. For all vehicles, the \emph{action} is \emph{acceleration}
$\in [-5, 4] \, m/s^2$.

\subsection{State Variables}

The three-vehicle setup considers nine states with different combinations of state variables used in the merge vehicle and traffic vehicle 
(see Table~\ref{tab:States} for details). The state set for the ego merge vehicle consists of a pair of variables (closing gap and closing speed), repeated several times for different objects: the traffic vehicle behind, the traffic vehicle in front, and the goal position. \citet{Claus1998} show that including the action in the state can improve performance. However, it is left to future work because~\citet{Schester2021} suggest the performance gain is marginal; possibly due to some innate encoding already present in the states.

The states always account for one traffic vehicle in the front and one in the rear. The speed and gap states default to a max gap of $100 m$ at zero closing velocity when a front or rear vehicle does not exist. The default $100 m$ max gap for a missing vehicle happens in training and testing. The reactive policy traffic vehicle uses a set of state parameters that are mostly different from those used for the ego merge vehicle. 
The traffic vehicles use the same set of parameters as the two-vehicle variant in~\citet{Schester2021}, except for an additional \emph{time in-between vehicle (TIV)} parameter corresponding to the gap from the rear to the front vehicle, divided by the velocity of the rear vehicle.

The algorithm clips out of range values to their limits; e.g, if "closing gap" is at $50 m$, the NN learns this as a $30 m$ value due to the clipping. A $30 m$ gap is roughly six typical car lengths ($\approx 1.5$ semi-truck \& trailers lengths). Therefore, longer distances are inconsequential. Figs.~\ref{fig:3vstrea} \& ~\ref{fig:4vstcon} (position differential results) support this assumption too. Other state ranges are determined similarly.

\subsection{Reward Function}
\label{sec:reward}

As in~\citet{Schester2021}, the general approach to the reward function is to give minor penalties for acceleration and deceleration, a larger reward for a successful merge, and a much larger penalty magnitude for collisions. The more severe penalties intend to evoke a stronger learned response by the algorithm because the expectation is that automated driving should be safer than human performance and as close to zero collisions as possible. Specifically, the reward for a successful merge is $10^3$, at-fault collision is $-10^5$, collision without fault is $-10^6$, and any acceleration or deceleration action is penalized by the magnitude of its value. 

\subsection{Training}

\begin{figure}
  \centering
  \includegraphics[width=\linewidth]{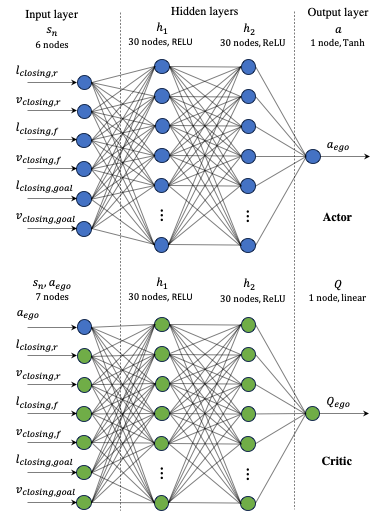}
  \caption{Three-vehicle actor-critic NN diagrams.}
  \Description{Three-vehicle actor-critic neural network diagrams.}
  \label{fig:3vehNN}
\end{figure}

A \emph{deep deterministic policy gradient (DDPG)} RL algorithm trains the merge vehicle NN in the three-vehicle setup. Fig.~\ref{fig:3vehNN} shows the NN diagram. DRL NN training occurs at each step of every episode. There are roughly $50$ steps on average per episode. The NN trains to $2.5M$ episodes, but the best performer is at $350K$, or about $17.5M$ training steps. However, the reward assignment for a collision or successful merge occurs only once per episode.

The input to the actor-critic DDPG NNs are the vehicle state variables. The output is the acceleration action. The NNs have two hidden layers with a $30$-node width and use rectified-linear and hyperbolic tangent for the activation function of the hidden- and output-layer units, respectively. The NN architecture for the reactive traffic vehicle is the same as for the Ego network, but not their parameters. 
The NNs train in parallel with a learning rate $=0.001$ (default) for both the actor and the critic, and the discount factor $\gamma=0.9$. The NNs use a memory replay buffer with batch size $=32$ and memory capacity $=10K$ samples. This architecture and parameter settings yielded reasonably good performance; yet it may benefit from further model selection.

The two vehicles in the traffic lane also have a TIV
 which is set uniformly at random from $[0.5, 2.5] \, s$ during training, but fixed at $0.8s$ during testing. For reference, ACC systems at highway speeds have at least a $1.6s$ min TIV. A study of driver behavior shows that most drivers prefer a min gap 
$\geq 0.6s$, and 
$< 3$\% maintain a gap 
$< 0.5s$ \citep{Nowakowski2011}. Hence, a $0.8s$ TIV is an aggressive and challenging yet reasonable and realistic test setting.

Each NN controller learns a policy mapping states (i.e., state parameter values) to actions (i.e., acceleration), both continuous in our model. 
The training uses a mixture of all three traffic action policies: \emph{random, constant, and reactive}. Independently from the merge vehicle, the two traffic vehicles select one of the three action policies at the beginning of each episode, and each traffic vehicle maintains that policy throughout the episode. The reactive policy of a traffic vehicle, selected $\approx 1/3$ of the time, also uses DRL to learn to avoid collisions and train the same NN controller. Therefore, the reactive traffic NN trains with $\approx 1/3$ times less data than the ego merge vehicle, so there is a mismatch in the amount of training experience between the two vehicle NNs.

\begin{figure}
\centering
    \includegraphics[width=\linewidth]{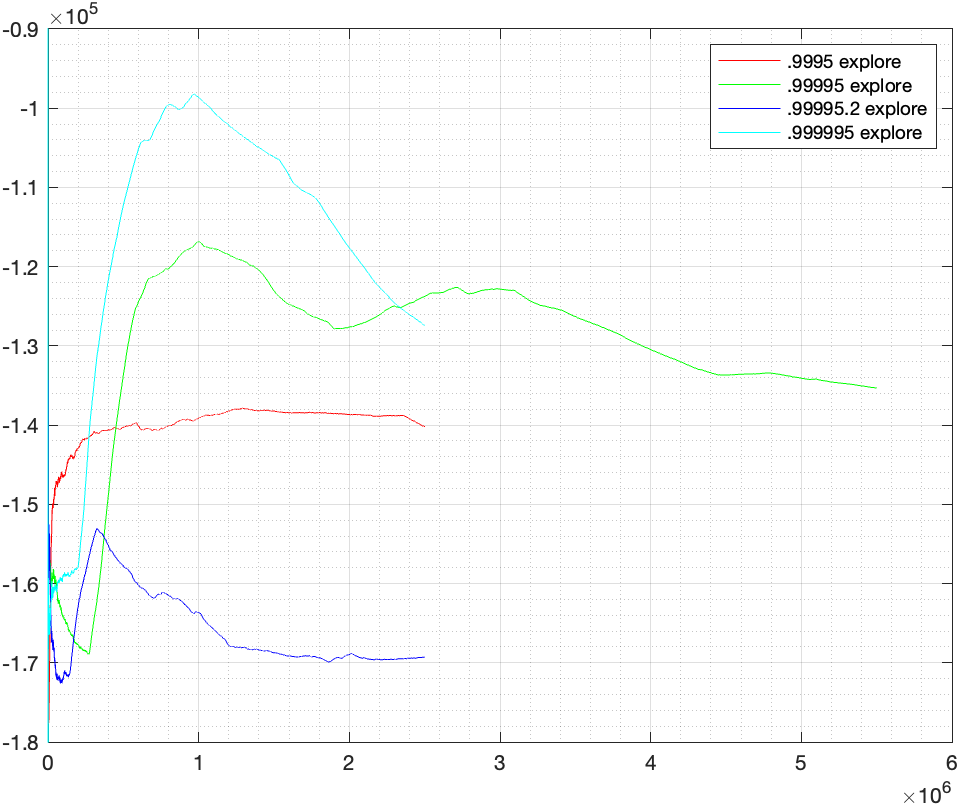}
\hfill
    \includegraphics[width=\linewidth]{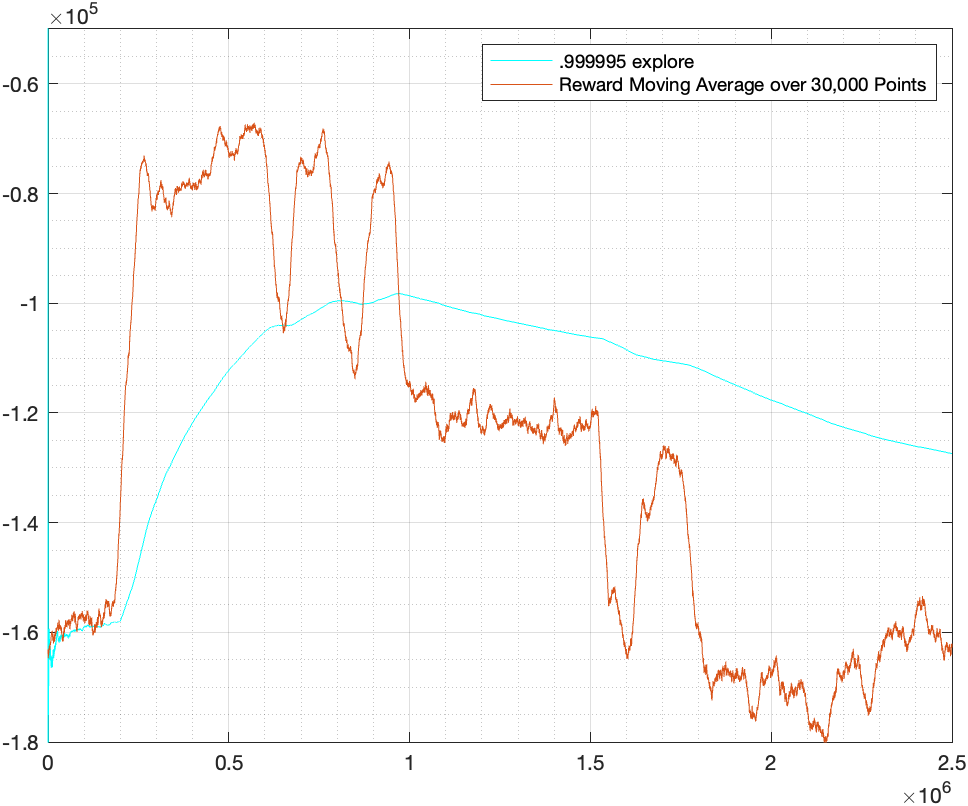}
\caption{Top: Cumulative reward average of four separate exploration values. Bottom: Cumulative reward average and moving mean for best variant: $0.999995$ explore.}
\Description{Plots of reward values for network training.}
\label{fig:3VehTrain}
\end{figure}

Exploration has strong effect on training. Just as in \citet{Lillicrap2015}, exploration is a noise value added to the action that decays exponentially at each iteration of each episode. Fig.~\ref{fig:3VehTrain} compares training runs with different exploration values: $0.9995$, $0.99995$, and $0.999995$. An exploration value of $0.999995$ is the best performer for the three-vehicle simulation (the curve with the highest peak).
(Cross-validating the exploration parameter is left for future work.)
The highest peak represents the cumulative average of reward values for all episodes trained. The curve selected has a peak of $-0.98E5$, around $!M$ training episodes. This peak is substantially higher than the other training trials.

It is common to see high variability in training performance between runs. Notably, two training runs in Fig.~\ref{fig:3VehTrain} use an exploration value of $0.99995$ (green and dark blue curves) and the same variables and settings, but the NN training randomly chooses values for start position, goal position, traffic vehicle length, TIV, speed, traffic action policy, initial (NN) controller parameters, among other factors. This variability creates challenges in determining the best overall performer or optimal configurations. However, this brute-force hand tuning and laborious art of evaluation is typical of NN tuning. Fig.~\ref{fig:3VehTrain} shows the cumulative average and the moving mean of the selected $0.999995$ training run. The cumulative peak value is around $1M$ episodes, but the highest average reward values are between $250K-950K$ episodes. The best network is at $350K$ episodes.

Fig.~\ref{fig:3VehTrain} also shows that good performance is fleeting. When training reaches a cumulative average peak, it is typical for the cumulative average to worsen. The blue and green curves prominently show the worsening behavior. While it is unclear exactly why this happens, this behavior is typical in NN training. A repercussion of this in the AD use case is that training would not be acceptable to continue indefinitely as it could lead to unexpected and poor performance. This idea is not intuitive for those who consider AI to be like human behavior because the common expectation is that once humans learn to merge into traffic well, they do not progressively get worse at it.

\begin{figure}
  \centering
  \includegraphics[width=1.0\linewidth]{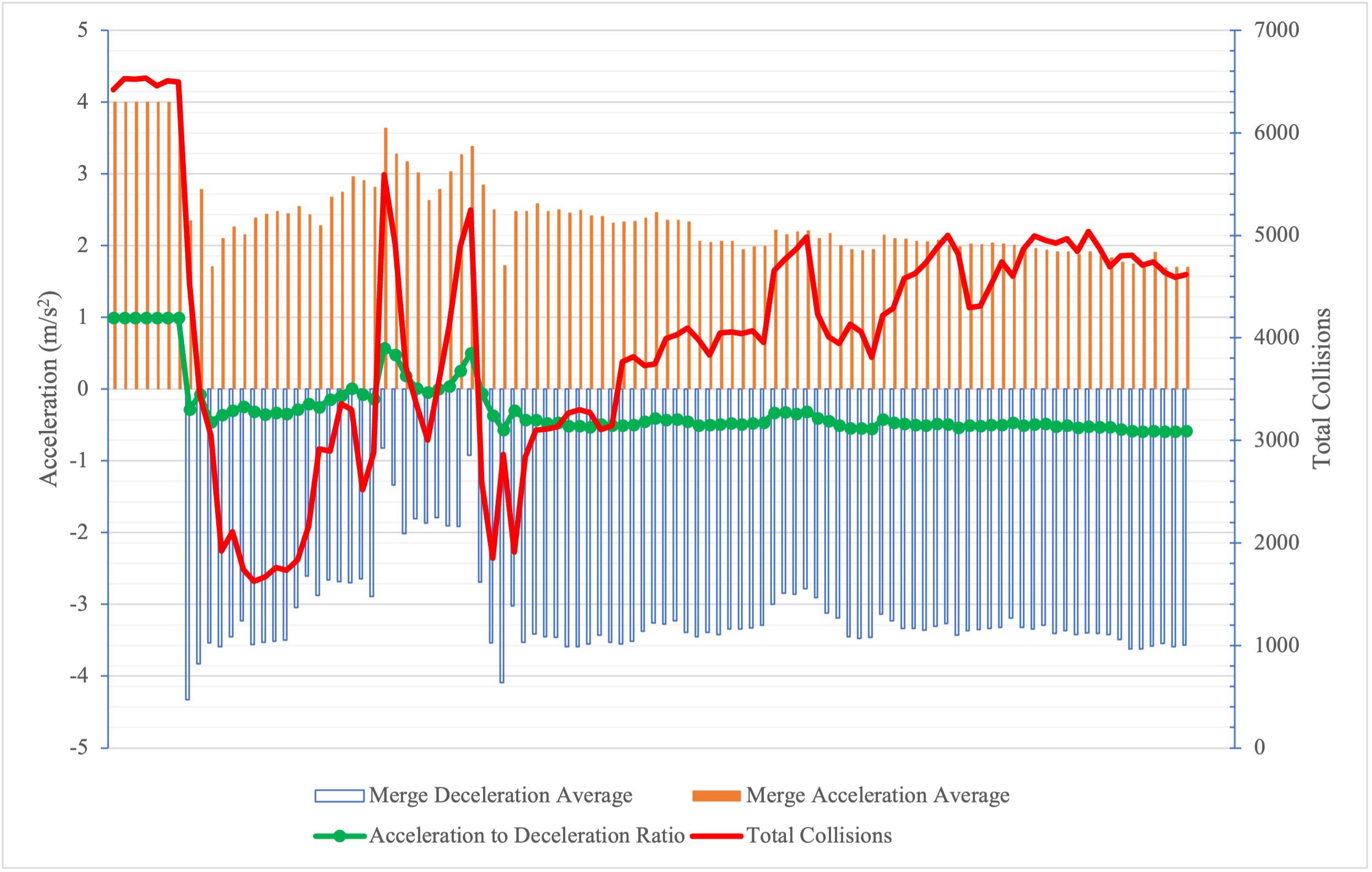}
  \caption{{\bf Standard test performance of the three-vehicle scenario with data sorted by training order.} Blue and orange bars indicate average deceleration and acceleration for each test instance. The green line is the difference between the acceleration and deceleration occurrence. The red line is the secondary axis (right), representing total collisions. Horizontal axis values indicate the tests at $25K$ intervals.}
  \Description{Bar chart and line representation of acceleration and deceleration values for training.}
  \label{fig:3vehmulplot}
\end{figure}

\begin{figure*}
  \centering
  \includegraphics[width=\linewidth]{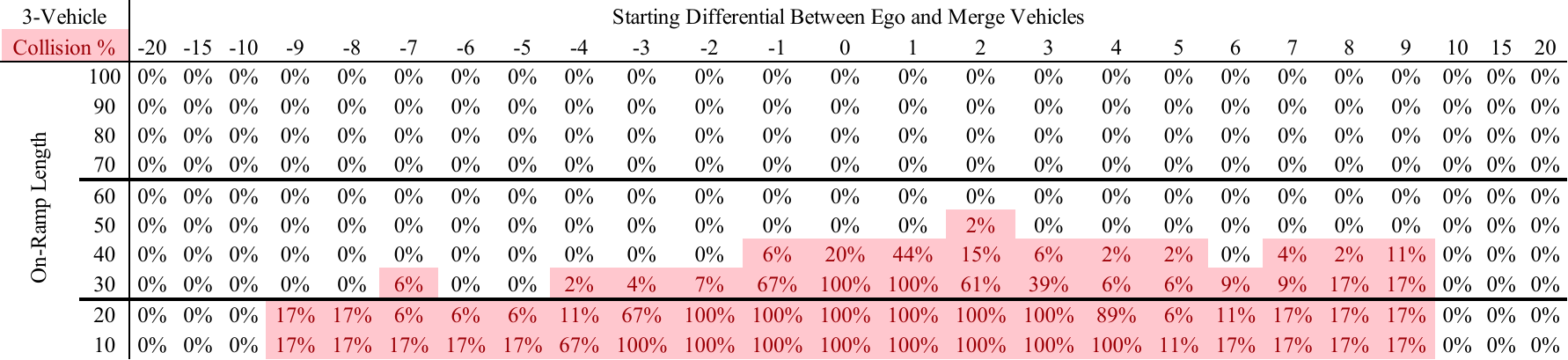}
  \caption{Three-vehicle collision test table. The vertical axis is the ramp length. The horizontal axis shows the starting position differential between the merge vehicle and the closest traffic vehicle. The percentages represent the frequency of collisions when testing against all three traffic actions policies of constant speed, random, and reactive. The MUTCD~\cite{AchmadAliFikriSyamsulArifin2022} suggests minimum taper lengths of $30 m$ and $60 m$ in urban and rural areas, respectively.}
  \Description{Table of testing results showing collisions between the vehicles in different lanes.}
  \label{fig:3vstrea}
\end{figure*}

\subsection{Best Controller Selection}

A MATLAB script selects the best network by summarizing relevant data from the saved test files created periodically during training. The script runs every $25$K episodes while training, saving network weights, and generating an output file that runs a test routine to generate performance evaluation data. Fig.~\ref{fig:3vehmulplot} is a multi-dimensional plot of the evaluation data. The best performing tests show a bias towards deceleration occurrences. Initially, the NN only selects action values at the acceleration limits of $4m/s^2$. When it beings to select both acceleration and deceleration actions, collisions rapidly decreases with a min at $350K$ episodes. Training continues afterwards, but the performance becomes erratic with total collisions shifting between high and low as episodes increase. There are two peaks of high total collisions around the $25$th and $34$th tests. These peaks correlate with greater acceleration bias and higher average acceleration values as well as lower average deceleration values. In addition, the slightly increasing bias towards greater deceleration starting around the $49$th test group corresponds to an increase in total collisions, the same as the decreasing performance of Fig.~\ref{fig:3VehTrain}.


The study of effective, automatic model-selection mechanisms in this setting is left for future work.

\subsection{Standard Test Results}

\begin{figure}
  \centering
  \includegraphics[width=1.0\linewidth]{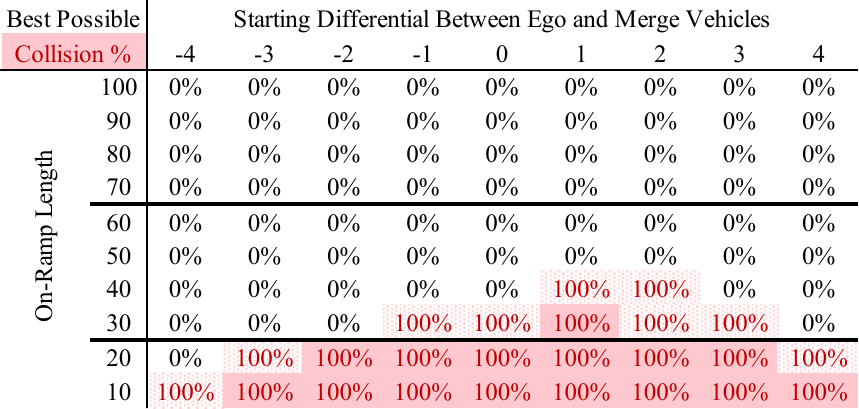}
  \caption{Ideal collision table of best possible performance showing on-ramp length and starting differential combinations. Collisions are unavoidable in the marked cells when the merge and traffic vehicles are acting using the most ideal action selection to avoid a collision. More lightly shaded cells show merge vehicle collision performance against a vehicle that maintains constant steady-state speed throughout the simulation.}
  \Description{Table showing collision results for an idealistic setting.}
  \label{fig:ideal_coll}
\end{figure}

Fig.~\ref{fig:ideal_coll} shows the best possible collision performance using idealized vehicle action settings while holding variables like initial speed and vehicle length constant. Fig.~\ref{fig:3vstrea} shows results tables at various gap settings from $5 m$ initial traffic gap to $100 m$. A $5 m$ gap is the same size as the test vehicles, making it a challenging condition because the ego vehicle needs to maneuver itself into a gap size the same as its length. A gap size of $10 m$ is also small because most drivers prefer to maintain a gap of at least three car lengths, which is $\approx 15m$~\citep{Nowakowski2011}. Gap sizes $\geq 15 m$ produce fewer collisions since the ego vehicle basically has only one traffic vehicle to contend with because the other vehicle begins to be far enough away to diminish its influence. While training is conducted with an equal mixture of random, constant, and reactive traffic action policies, the two traffic vehicles use the same policy (either constant, reactive, or random) during each standard test episode. This evaluates performance in a more standard way versus dissimilarities in action between the two traffic vehicles with mixed policies. The results table is generated using a mixture of all three policies to represent performance against all types of traffic vehicle behaviors. The ego merge and traffic vehicles have clearly learned behavior that allows them to avoid collisions in most instances of starting gap deltas and goal positions.

\section{Full Scene Simulation}

The Full-Vehicle Scene 
(Fig.~\ref{fig:4vehsim})
consists of two vehicles in the merge lane and two or more in the traffic lane. This merge scene is representative of a real-world traffic scene. Game-theoretically, the strategic landscape is considerably more complex. Additional vehicles both in the traffic or merging lanes can drastically alter or constraint an optimal controller's solution space because "obvious" policies (fully accelerate or decelerate) may not be viable to the merging vehicle; e.g., a second traffic-lane vehicle could facilitate or impede merging, and a merging-lane vehicle in front drastically restrict acceleration. It terms of scope \& complexity, the modeling challenge in moving to more than two vehicles is the potential exponential dependency on the number of vehicles in terms of states \& actions, like those from most naive/obvious extensions. We address this challenge by proposing a modeling approach to the full scene is "reductionist," yet complete. The full scene model is also dynamic, by identifying and adapting to the closest/most-relevant vehicles, allowing for great flexibility in its application, thus increasing its generalization ability to arbitrarily-sized scenes, while avoiding the exponential dependency.

In this version, the traffic vehicles remain in the scene once the rear ego vehicle moves ahead or behind the two-vehicle front-rear traffic pair, compared to the three-vehicle simulation which kept only the two closest traffic vehicles. Overall, the results for the full scene are similar to the three-vehicle results.

The rear merge vehicle is the AD ego vehicle, and the front merge vehicle trains and uses the same NN as the rear ego vehicle. The merge vehicle in front represents the first vehicle in front of ego. There could be more than one vehicle in front of ego, but ego’s immediate in-lane concern will always be the vehicle in front of it. The front merge vehicle constraints the acceleration behavior allowed, and ego needs to react if the front vehicle slows. The two traffic vehicles are always in front of and behind the ego vehicle. This setup is a condition of the training and testing simulation for the ego vehicle. If the front merge vehicle is in front of the front traffic vehicle, it assumes another vehicle is ahead of it at $100 m$ at $0 m/s$ closing speed. These default placeholder values train the DRL network to always assume that there is a traffic vehicle in front and the rear, even if it is very far away. 

The state set for the full scene is similar to the three-vehicle setup. 
Table~\ref{tab:States} shows the state variables. The merge vehicle states remain the same, except for an added TIV variable to the front merge vehicle. The traffic vehicle states remain very similar too. They grow from a set of five states to a set of eight because of the additional merge vehicle closing gap, closing speed, and proximity (behind or in front). 

\begin{figure*}
  \centering
  \includegraphics[width=0.95\linewidth]{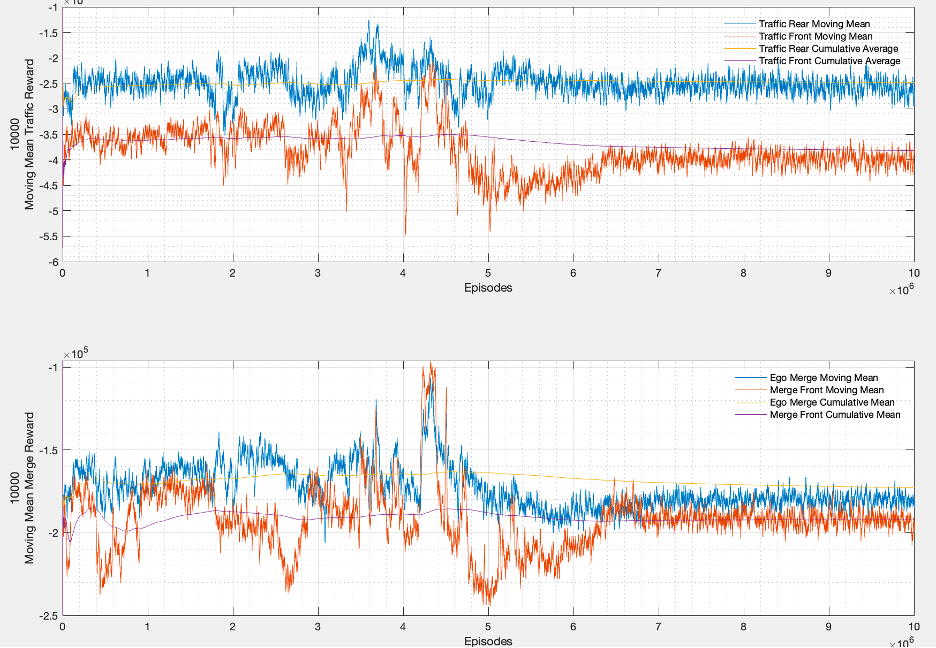}
  \caption{Training Graph of Moving Mean and Cum. Avg. for Rewards. Near $7M$ episodes, the acceleration action for both the merge and traffic vehicles settle to continuously choose an action limit value of $-5 m/s^2$ acceleration, regardless of state value.}
  \label{fig:4vehtrain}
  \Description{Plot of training data showing moving mean and cumulative average of rewards.}
\end{figure*}

\begin{figure*}
  \centering
  \includegraphics[width=\linewidth]{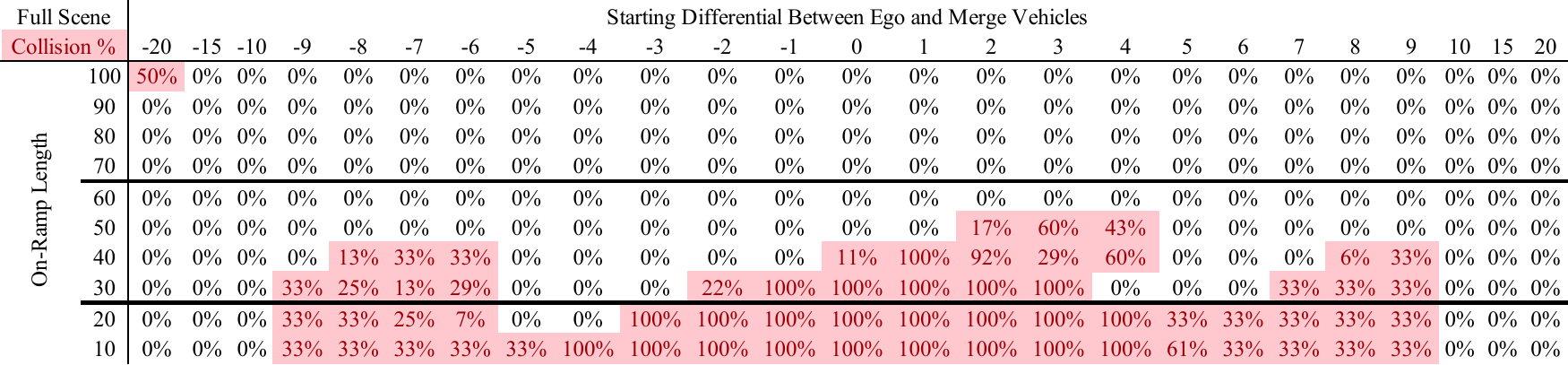}
  \caption{Full-Scene Collision Test Table. Constant and random policies standard test table with multiple gap settings of $5m$, $15m$, and $25m$.}
  \Description{Table of testing results for full-scene showing collisions between the vehicles in different lanes.}
  \label{fig:4vstcon}
\end{figure*}

\subsection{Controller Training and Performance}

Training is as in the three-vehicle scene. Both merge lane vehicles update the NN controller at each step of each episode. All traffic vehicles that have a reactive policy update the traffic NN controller. 
The merge and traffic NNs use different/separate weight parameters.

Fig.~\ref{fig:4vehtrain} shows the training graph of cumulative rewards and moving averages over $10$ data points for the front and rear merge and traffic vehicles. The graphs only show the front and rear vehicles, even if more are in the scene. Training runs to a total of $10M$ episodes. The peaks in the data for both the merge and traffic vehicles suggest that the best performance occurs between $4.2M$ \& $4.5M$ episodes. It shows training performance settle after about $7M$ episodes. This behavior is typical during training of the NN controllers. Early on in the training, the performance increases and decreases, but on the whole, it generally improves and peaks. After the peak, cumulative performance decreases until the reward value settles towards a repetitive value within the band of randomness for the training variation. The reward value settles because the NN learns to stick with an extreme value at the acceleration limit, either $-5m/s^2$ or $4m/s^2$. As training continues, the NN never learns to deviate from the extreme value limit. However, the ground truth requires both acceleration and deceleration values to achieve ideal performance. The only way to achieve ideal performance is to have both acceleration and deceleration actions. Therefore, the settling behavior that occurs is not a learned ideal performance. Fundamentally, it shows that the NN has reached a local minimum and is unable to break away from it.

The MATLAB performance testing script runs with the output files generated every $25K$ episodes during training. The best network is at $4.325M$ episodes with total collisions of $4,389$. This best performer at $4.325M$ episodes corresponds to the peak in the moving average graph of Figure~\ref{fig:4vehtrain}.

\subsection{Standard Test Results}

Figure~\ref{fig:4vstcon} shows the combined random and constant policy test results for the full scene. The random policy selects an action in $[-5,4] m/s^2$ at each $100 ms$ time step of the simulation. The constant policy keeps a constant speed for the traffic vehicles during each episode. The speed only decreases when a rear traffic vehicle is below the min TIV threshold of $0.8s$. Maximum deceleration action of $-5m/s^2$ continues at each time step until the TIV reaches its threshold. If the TIV is $\geq 0.8s$, the traffic vehicle acts with zero acceleration and maintains speed. 

We use gap settings of $5m$, $15m$, and $25m$ to generate the full range of ramp length and initial differential position combinations. A gap setting of $5m$ is a very challenging gap setting as mentioned previously. The collisions are fewer as the initial position reaches $5m$ or $-5m$ but increases as the starting differential initial positions reaches $10m$ or $-10m$. Taking action to avoid a collision is easier at $5m$ differential because the end of the merge vehicle aligns with the end of the traffic vehicle. However, because the traffic vehicles regenerate in the simulation, it becomes difficult to avoid collisions again.

As the initial gap increases to $\geq 15m$, the results are considerably better. In fact, at $15m$ they have essentially achieved optimal performance. These results show that optimal performance is still achieved despite the additional vehicles in the busy road scene. Edge cases still exist despite the good overall performance. For example, the $25 m$ initial gap test case shows near-optimal performance, similar to the $15 m$ setting. However, there is a collision at the $100 m$ goal position and differential starting position of $-20 m$.

\section{Future Work}

The evaluation settings are fixed in many ways, but future work could evaluate different initial speeds, combinations of traffic reaction policies, or other variables. The simulator assumes that no additional lanes exist beyond the first lane of traffic, but it is possible at times for the traffic vehicle to change lanes to avoid a collision with the merge vehicle; combining the study of lane merging with lane changing. Steering control and in-lane maneuvering could be added to the state set to understand if more complex vehicle control is beneficial. As discussed earlier, improved performance is shown when the action is included as part of the state. Acceleration can readily be captured by modern vehicle sensors, so future work should compare and contrast its inclusion in both the merge and traffic vehicle state sets.

The work presented did not attempt to fully optimize model selection and training, which may yield significance performance improvements.
For example, the settling of the NN at action limit extremes suggests a need for network optimization since both limit values are needed in the ideal case of Fig.~\ref{fig:ideal_coll}. Exploration values \& schemes, learning rates, batch size, automated network selection \& recursive training from known best networks, cross-validation, hidden layer count, layer width, activation functions, or other RL network approaches like LSTM ~\citep{10.1162/neco.1997.9.8.1735,10.1162/neco_a_01199} or A3C ~\citep{pmlr-v48-mniha16} are all understudied areas for this work. Adding noise during training to mimic real senor data and other imperfections could increase robustness. Similarly, expanding the limits of acceleration, yet training with under-powered vehicles would mimic actual scenarios that could be encountered. Training was all done using self-play, but inverse-RL from existing datasets could be useful, especially since it is unclear if human-level performance has been achieved in any on-ramp merging work.





\section{Contributions}

Two MA DRL merge models were presented in this work, an intermediate three-vehicle scene and a full merge model with two merge vehicles and at least two traffic vehicles. The three-vehicle model shows that it can avoid merging collisions well below the typical recommended ramp lengths of $256m$. Scaling to the full-scene continued to produce good results with near optimal performance in constant and random policies in reasonable gap settings $\geq 15m$. Yet, better model selection and training may yield further improvements. 
Ultimately, the results show viability in the approach of controlling the speed of the vehicle to avoid a merging collision. Future testing could show if this idealized approach can be implemented to control the speed in a more sophisticated real-world environment.

\begin{acks}
LEO was supported in part by NSF Award IIS-1907553.
\end{acks}



\bibliographystyle{ACM-Reference-Format} 
\bibliography{AAMAS_2025}


\end{document}